\documentclass[10pt,twocolumn,letterpaper]{article}

\usepackage{cvpr}
\usepackage{times}
\usepackage{epsfig}
\usepackage{graphicx}
\usepackage{amsmath}
\usepackage{amssymb}
\usepackage{comment}

\usepackage[pagebackref=true,breaklinks=true,letterpaper=true,colorlinks,bookmarks=false]{hyperref}

\usepackage{lipsum}

\newcommand\blfootnote[1]{%
  \begingroup
  \renewcommand\thefootnote{}\footnote{#1}%
  \addtocounter{footnote}{-1}%
  \endgroup
}

\cvprfinalcopy 

\newcommand{\RGBnet}{RGBPolicyNet}
\newcommand{\StructureNet}{ObjectPolicyNet}

\def\cvprPaperID{2290} 

\ifcvprfinal\fi
\begin{document}



\title{Reward Learning  from Narrated Demonstrations}



\author{Hsiao-Yu Fish Tung \phantom{1235}  Adam W. Harley \phantom{1235}  Liang-Kang Huang  \phantom{1235} Katerina Fragkiadaki  \\
Carnegie Mellon University\\
5000 Forbes Ave, Pittsburgh, PA 15213\\
{\tt\small \{htung, aharley, katef\}@cs.cmu.edu, henrykang7177@gmail.com } 
}

\maketitle

 \blfootnote{The work has been accepted to Conference on Computer Vision and Pattern Recognition (CVPR) 2018.}
\begin{abstract}

Humans effortlessly ``program" one another by communicating goals and desires in natural language. 
In contrast, humans program  robotic behaviours by indicating desired object locations and poses to be achieved \cite{hindside}, by providing  RGB images of  goal configurations \cite{DBLP:journals/corr/FinnL16}, or supplying a demonstration to be imitated \cite{DBLP:journals/corr/DuanASHSSAZ17}. None of these methods 
generalize across environment variations, and
they convey the goal in awkward technical terms.
This work proposes  
 joint learning of natural language grounding and instructable behavioural policies reinforced by  perceptual detectors of natural language expressions, grounded to the sensory inputs of the robotic agent.
 
Our supervision is narrated visual demonstrations (NVD), which are visual demonstrations paired with verbal narration (as opposed to being silent). 
We introduce a dataset of NVD where teachers perform activities while describing them in detail.  
We map the teachers' descriptions to perceptual reward detectors, and use them to train corresponding behavioural policies in simulation. 
We empirically show that our instructable agents (i) learn visual reward detectors using a small number of examples by exploiting hard negative mined configurations from demonstration dynamics,  (ii)  develop  pick-and-place policies using  learned visual reward detectors, (iii)  benefit from   object-factorized  state representations that mimic the syntactic structure of natural language goal expressions, and (iv) can execute  behaviours that involve novel objects in novel locations at test time, instructed by natural language.  
 
 \end{abstract}

\section{Introduction}

\begin{figure}[t!]
    \centering
    \includegraphics[width=1.0\linewidth]{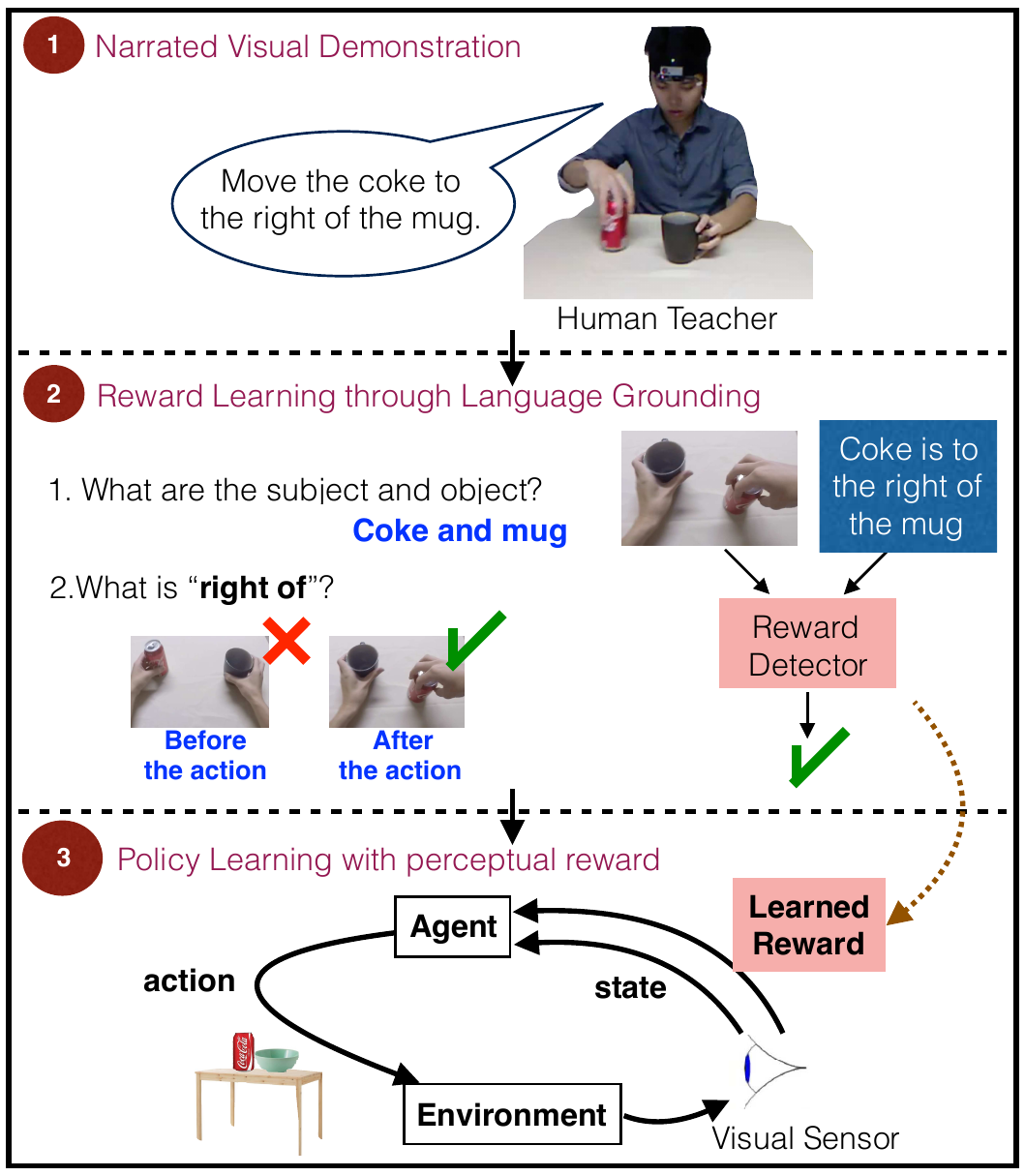}
     \centering
        \caption{\textbf{Reward Learning  from Narrated Demonstrations}. We begin with a narrated visual demonstration, prepared by a human (1). Our system then learns a spatial relationship detector from the visuals and audio (2). Finally, we use the learned detectors to train pick-and-place policies (3).
        }
    \label{fig:pipeline}
\end{figure}

\begin{figure*}[t!]
    \centering
    \includegraphics[width=1.0\linewidth]{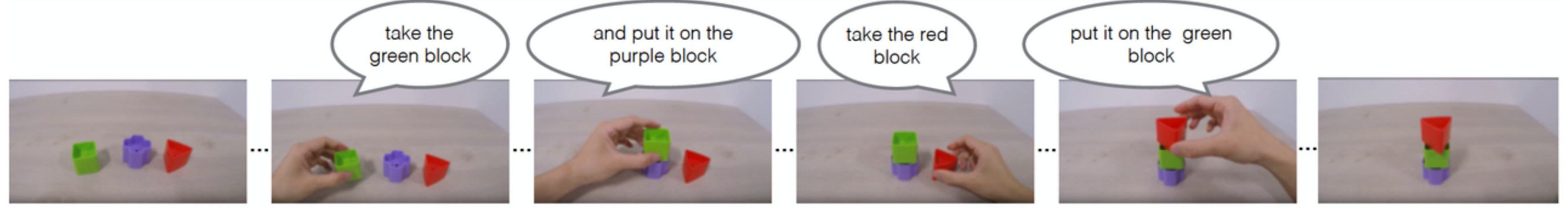}
     \centering
        \caption{\textbf{Narrated visual demonstrations}. The teacher  demonstrates activities and concurrently narrates them in natural language using a microphone.   Many related tasks are demonstrated densely  in time;   
        temporal segmentation of the demonstration video into different tasks is  easy based on natural language sentences. }
    \label{fig:dataset}
\end{figure*}

Currently, rewards or goals for behavioural policy learning are either manually coded by experts \cite{DBLP:journals/corr/abs-1709-10087, DBLP:journals/corr/LevineFDA15,lenz_deepMPC_2015},  
or are learned from human supplied demonstrations (LfD, or inverse RL) \cite{MullingKKP2012_2,Ng:2000:AIR:645529.657801,DBLP:journals/corr/FinnLA16,ziebart2008maximum}. 
Manually coded rewards are hard to generalize across variations of the environment. Moreover, we often need a large number of demonstrations for the right reward function to be effectively communicated to the agent,  
invariant to 
distractors, accidental coincidences, view-dependent feature representations, speed of execution, etc. In contrast, humans  effortlessly program each other's behaviour \textbf{by conveying goals and desires in natural language}, e.g. ``for a CVPR submission, the margin should be one inch on each side of the page", or ``while driving, make sure  to keep a safe distance from the car in front of you."  
Interestingly, in absence of natural language competence, \textit{understanding the goal of a behaviour is often harder than learning the behaviour itself.}
For example, although macaques are excellent tree climbers, incrementally
training them to pick
coconuts (with RL) is extremely laborious \cite{LupBer15}.
Humans, on the other hand, can easily understand the goal of ``picking coconuts", but are less capable of carrying it out. 

This work introduces   \textbf{instructable perceptual rewards}, namely, reward functions that can be both expressed in natural language and detected in the visual sensory input of the agent. It further proposes a framework for learning these rewards from   \textbf{Narrated Visual Demonstrations} (NVD), which are visual demonstrations synchronized with natural language descriptions. 
Rather than struggling to discover essential goals of human behaviour from a large number of \textit{silent} visual demonstrations, we instead consider \textit{narrated} visual demonstrations, where narrations 
describe actions being taken, objects involved, and goals achieved, as shown in Figure \ref{fig:dataset}. 
Given a set of NVDs, we first learn to ground natural language utterances that express activity goals---rewards,  \eg, ``coca cola on top of the book," to modular neural visual detectors. We then use such visual  detectors to reinforcement learn  policies that achieve the corresponding goals (see also Figure \ref{fig:pipeline}). 

Narrated visual demonstrations are more data-efficient than their silent counterparts. We empirically show that learning instructable perceptual rewards and corresponding policies from NVDs results in data reduction for both reward and policy learning.  
This reduction comes from (i) leveraging large-scale annotated static image datasets \cite{krishnavisualgenome} of objects and visual  relationships to help ground natural language goal descriptions, (ii) demonstration dynamics,  
where similar objects
 appear with different attributes or relationships in consecutive demonstrated tasks, which forces our reward detectors to focus on the \textbf{temporal transformation of such arrangements/attributes}, as opposed
to  object detection and recognition, and, (iii)  object-factorized state representations as input to reward detectors and policy networks, mimicking the syntax of natural language descriptions.  

Collecting narrated visual demonstrations is scalable. We  collect a dataset of pick-and-place activities using cameras and microphones mounted  on demonstrators (human teachers) that perform activities while verbally narrating them (see Figure \ref{fig:dataset}).  Automated speech recognizers map the narrations to transcripts temporally synchronized with the visual demonstrations.  Each video contains multiple, diverse demonstrations, proceeding one another closely in time. Temporal segmentation of  sequential demonstrations \cite{DBLP:journals/corr/HausmanCSSL17} is easily obtained by considering the segmentation of the transcript into verbal phrases;  this alleviates the current need for demonstrations to  concern a single isolated task at a time \cite{DBLP:journals/corr/HausmanCSSL17}. 
In terms of detail, deliberate demonstration and verbal narration is  more scalable than post-hoc captioning \cite{rohrbach2012database}, and allows natural language descriptions that are dense in time, without overwhelming the demonstrator. 
The videos in our 
dataset 
are instructional in nature, similar to instructional videos on YouTube \cite{DBLP:journals/corr/HuangLLN17,Alayrac16unsupervised}. While Youtube videos target on audience with advanced language  grounding capabilities, our dataset instead attempts to teach such natural language grounding, alongside the demonstrated behaviours.   
To the best of our knowledge, no previous work hasconsidered narrated videos for learning  rewards and policies for the demonstrated actions.

In summary, our contributions are:
    \begin{itemize}
       \item We introduce instructable perceptual rewards as modular visual detectors of natural language expressions of activity goals, and show how to learn them from few NVDs by exploiting demonstration dynamics for effective hard negative mining. 
       
       \item We   introduce a  dataset of NVDs of daily activities. We show that pairing visual demonstration with natural language narration permits scaling up the collection of visual demonstrations,  which can now be dense in time and  depict diverse tasks, as opposed to being structured, isolated in time, and depicting a single task. 
       
              
                     \item We demonstate that our agent effectively learns instructable policies using \textit{noisy} instructable perceptual reward detectors, and can execute novel behaviours at test time, exploiting compositionality of natural language.  
        
        \item We show that object-factorized state representations for our policy network generalize better than frame-centric RGB input.  
              

    \end{itemize}

\section{Related Work}

In the absence of a manipulation language, previous works convey goals using  RGB images \cite{DBLP:journals/corr/FinnL16},  demonstrating the desired activity itself \cite{DBLP:journals/corr/DuanASHSSAZ17},  supplying desired 3D poses of the objects and end-effectors in a particular scene \cite{DBLP:journals/corr/AndrychowiczWRS17}, or assuming that there is only one behaviour that can be requested \cite{45942}. This work proposes expressing goals in natural language, and builds corresponding perceptual detectors that can drive reinforcement learning for policies.

\paragraph{Mapping instructions to actions}
Numerous works have proposed learning a mapping from instructions to high level  action sequences of the agent. 
For instance, paired examples of instruction and action sequences have been collected through Amazon Mechanical Turk \cite{DBLP:journals/tacl/ArtziZ13,Misra-RSS-14,Misra2015EnvironmentDrivenLI,tellex11}. 
Other works attempt to learn such mapping using reinforcement learning, from pairs of instructions with desired final goal configurations \cite{Branavan:2009:RLM:1687878.1687892,DBLP:journals/corr/MisraLA17}. These models execute the predicted action sequences and evaluate whether the desired goal state is reached. 
Most approaches consider  action sequences to be given in the task space of the agent.  Instead,     we consider  third-person demonstrations and narrations, where the automated visual recognition needs to infer the locations of objects and their spatial configurations. We train our agent with reinforcement learning using perceptual detectors of the natural language goal expressions, as opposed to direct imitation of the corresponding action sequences. 


\paragraph{Visual imitation learning} 
Visual imitation learning (VIL) considers
the problem of acquiring skills by observing visual  demonstrations  \cite{DBLP:journals/corr/StadieAS17}.  
It  requires inference of the ``reward" (i.e., the goal of the behaviour) that the agent will attempt to match by self-practice, and adapting the demonstrations to its own degrees of freedom and workspace. 
Numerous works circumvent the difficult visual perception problem in VIL using special instrumentation of the environment to read off object and hand poses during video demonstrations  \cite{DBLP:journals/corr/KroemerS16}, or  use rewards based on \textbf{known goal 3D object configurations.} 
A notable exception is the work of Sermanet, Xu, and Levine \cite{DBLP:journals/corr/SermanetXL16}, which learned perceptual rewards for a pouring task, using a large number of visual demonstrations. 
In this work,
we instead propose 
\textbf{narrated} visual demonstrations for joint learning of natural language grounding and reward detectors.  
Natural language casts attention to the relevant parts of the video (e.g., the relevant objects), and facilitates the mapping of natural language descriptions to visual reward detectors. At test time, we can easily program novel behaviours  by composing novel natural language goal descriptions.  



\paragraph{Perceptually-grounded natural language}
Language grounding has recently attracted a lot of attention, with the introduction of large-scale image captioning, video summarization  \cite{rohrbach2013translating,fang2015captions} and visual question answering datasets \cite{antol2015vqa,MovieQA}.   
Captioning models describe images and videos using natural language sentences \cite{devlin2015language} and visual question answering models answer queries about an image \cite{andreas2016learning}. Such vision/language models are supervised by image captions or question/answer pairs collected from AMTurkers \cite{rohrbach2012database}, subtitles from movies \cite{rohrbach2017generating}, or movie descriptions for the blind \cite{rohrbach2015dataset}. This paper has an orthogonal goal to the aforementioned works: we are interested in learning to ground natural language descriptions of goal configurations to visual input, and use this mapping as a reward detector for policy search. \textbf{This replaces manually-coded rewards with natural language instructions.}


\paragraph{Object-factorized state representations}
The recent work of Kansky et al. \cite{DBLP:journals/corr/KanskySMELLDSPG17} showed how object-factorized state representations and dynamics can generalize across environmental variations, in contrast to frame-centric policies. In that work, it was assumed that the object identities were known beforehand. Here, we use natural language as weak supervision to focus attention to relevant objects, and use object detectors to learn reward configurations, and also during policy training and testing, to supply object-factorized states as input to out policy network.  Other works \cite{DBLP:journals/corr/FragkiadakiALM15,DBLP:journals/corr/BattagliaPLRK16} have considered object-centric predictive models of motion under close-by interactions, and showed they generalize better than frame-centric models.


\section{Instructable Reward and Policy Learning from Narrated Visual Demonstrations}

\subsection{Collecting Narrated Visual Demonstrations} \label{sec:dataset}
We collect narrated visual demonstrations using GoPro cameras and  microphones mounted on the head of the human demonstrator.  The demonstrator names objects in the scene, describes their relationships, indicates the activities  performed, explains the outcomes, and gesticulates deliberately so as to guide the learner towards the correct  interpretation of the  natural language description. Verbal narrations are automatically transcribed into textual descriptions using the Google speech recognition API \cite{googlespeech}. Mistakes of the speech recognizer are rare and are corrected by hand. The sync of the narration to the video, along with the present-tense descriptions, provide a natural alignment of the semantic content to the visual stream, e.g., \textit{``I am placing the cup on the opening of the bottle"}. Consecutive demonstrations are temporally segmented using their alignment to natural language utterances. This convenient segmentation method is only possible with narrated (rather than silent) demonstrations. In terms of human effort, the scalability of verbal narrations far surpasses annotation methods considered in previous works, such as video post-transcription \cite{rohrbach2012database} or scene graph annotations \cite{krishnavisualgenome}. 
Each video is between three and five minutes long and contains 14 to 30 individual demonstrations of short  activities. 
We have thus far collected two hours of densely annotated videos. This paper uses the pick-and-place activities of the dataset (around 10 minutes in total) to learn reward detectors and train corresponding pick-and-place policies in simulation. Many more diverse activities are contained in the dataset, which we will make publicly available.  We are not aware of a dataset of paired videos and natural language descriptions that addresses  natural language grounding for skill policy learning, which is a gap our work attempts to cover.

\subsection{Learning  Instructable Perceptual Rewards through Natural Language Grounding} \label{sec:rewards}


We learn visual reward detectors by grounding natural language descriptions of goals of pick-and-place activities (\eg, ``the coke can is on top of the book") to  modular neural programs that take an image and description as input, and output a score of how well the image matches the description. These reward reward detectors are used to train pick-and-place policies to achieve the configuration instructed by the natural language expression.


  
\begin{figure*}[h!]
    \centering
    \includegraphics[width=1.0\linewidth]{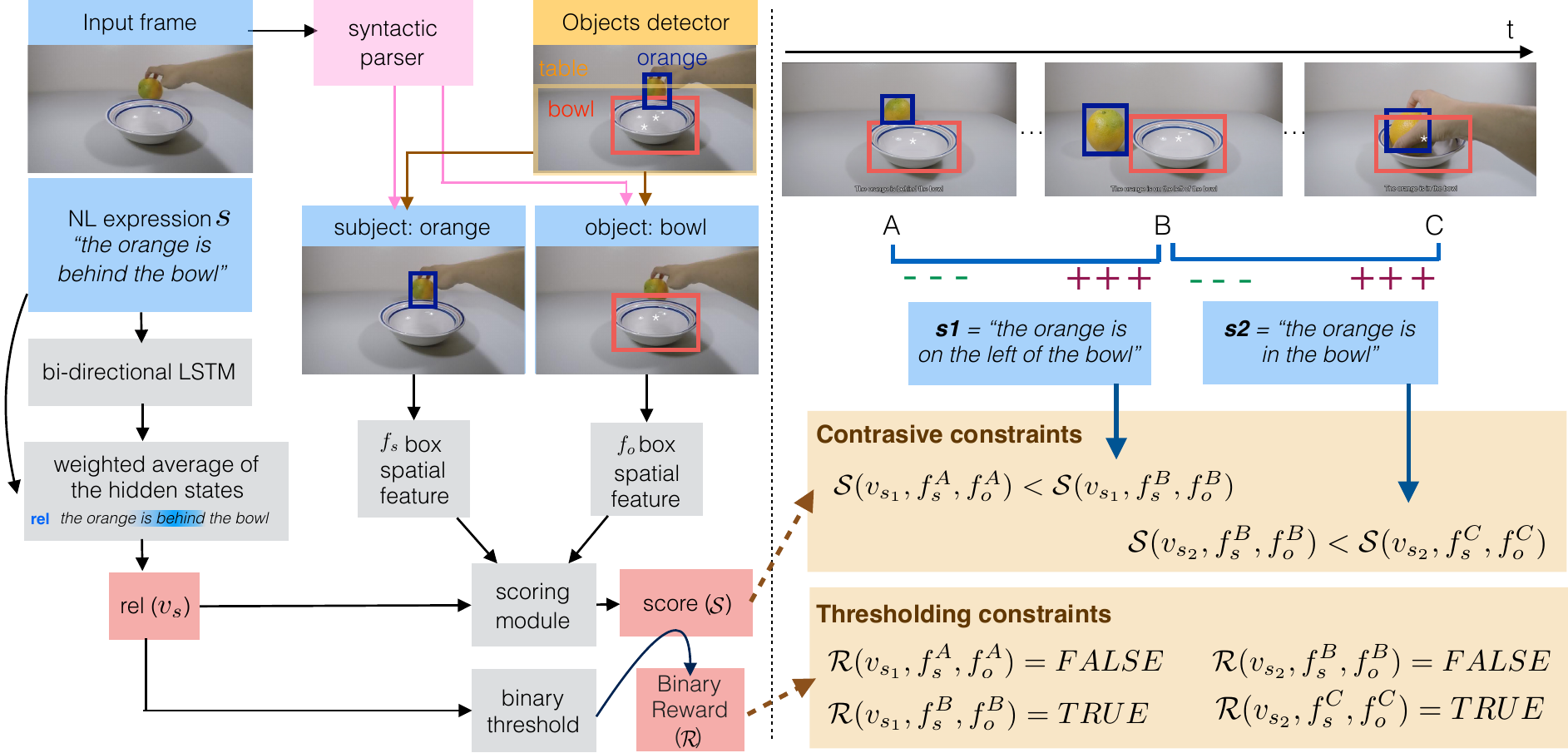}
     \centering
        \caption{\textit{Left:} \textbf{Visual detection of natural language spatial expressions} comprised of two object detectors and a relation module that computes a relation word embedding vector and given the spatial features extracted from the detected boxes, outputs a score for the spatial configuration. The score can be further transformed into
        a binary reward using predicted
        threshold. 
        \textit{Right:} \textbf{Hard negative mining} using spatial  configurations of adjacent in time video frames, demonstrating related natural language expressions. Related in time demonstrations provide hard negative examples to our relation module, free of static image biases, that allow it to improve from very few examples.    }
    \label{fig:referential}
\end{figure*}

Our visual detectors combine object detector modules and pairwise relation modules, assembled  based on the syntactic structure of the natural language description, provided by a syntactic parser \cite{chen2014fast}. 
The architecture is depicted in Figure~\ref{fig:referential}. 
It is comprised of two object detectors for the subject and object in the natural language expression,
and a relation neural module for scoring their spatial configuration. 

The object detectors build upon the state-of-the-art faster RCNN architecture \cite{DBLP:journals/corr/HuangRSZKFFWSG016}, and have been pretrained in Visual Genome \cite{krishnavisualgenome} and COCO datasets \cite{DBLP:journals/corr/ChenFLVGDZ15} to detect objects from 3000 categories. We use the Stanford syntactic parser \cite{chen2014fast} to parse the natural language expression into subject and object strings, and use the appropriate outputs of the object detectors to localize the mentioned object categories in the image. If the objects do not have a high enough detection score, we discard the corresponding frame.   
The relation module takes as input (i) a word embedding of the spatial relationship  $v_s$,  computed using a weighted average of the hidden states of a Bidirectional LSTM (BiLSTM)  over the natural language expression's words, where the weight distribution is predicted by the same BiLSTM, 
and, (ii) spatial locations of the object and subject, encoded as normalized pixel coordinates, and the width and height of the detected bounding boxes. This module outputs a score for the corresponding spatial relationship, as shown in Figure~\ref{fig:referential}-left. The relation module is pre-trained 
to localize referential expressions in the Visual Genome image dataset \cite{krishnavisualgenome}, as part of the model of \cite{modularreferential}. 
Although a referential expression, such as ``the orange in the bowl", is not identical in meaning to a description, such as  ``the orange is in the bowl", or to a desired post-condition, such as  ``the orange should be in the bowl", in practice their learned embeddings are similar. 

Our model is a variation of the referential expression detector of \cite{modularreferential}. The difference between the two is that, instead of object detectors, their model uses two localization modules, which, take as input a weighted average of the hidden state of the BiLSTM for the subject and object, and the visual features aggregated within a bounding box proposal, and score the probability that the bounding box proposal captures the referred subject or object, respectively.  
Their expression detector  sums the scores of the two localization modules and the relation module to score how well the two considered object proposals convey the referential expression. 
We instead use object detectors that take visual features as input, and predict an object category within a predefined set of categories. Linguistic variability  can be handled by considering the inner product of the word embedding of the detectable object categories with the  word embedding of the subject and object of the utterance, considered in the model of \cite{modularreferential}. Thanks to its modularity, the detector generalizes  better than a monolithic network trained to map a single frame or bounding box to spatial configurations scores. 

\paragraph{Weakly-supervised metric learning with hard negative mining}

In our NVD dataset, video frames are paired with corresponding pieces of  transcript, as generated by the speech recognizer. We temporally segment a  video  sequence into individual demonstrations whenever two consecutive natural language utterances are different. Our reward detector  is trained from such automatically aligned  utterance-frame pairs, the same utterance covers all frames of the demonstration. We consider only the frames paired with exactly one natural language utterance and   finetune  the relation module of our reward detector  using metric learning. Specifically, we ask our relation module to score higher in the frames paired with the considered utterance at the end of each demonstration (input to our relation module) and lower at the  frames in the beginning of the demonstration. 


Let $f_s^k$ and $f_o^k$ be the spatial features (normalized pixel coordinates, width and height) of the detected  boxes  for the subject and the object of the paired natural language utterance in frame $k$, respectively, and let $v_s$ denote the 
 relation embedding vector produced by the BiLSTM, given  natural language expression $s$. 
 For each video segment paired with natural language utterance $s$, let $X^-_s$ denotes the indices for the first few frames (negative examples) 
 and $X^+_s$ denotes the indices of the last few frames, (positive examples for the goal configuration), as shown in Figure  \ref{fig:referential} \textit{right}. 
Then, the contrastive loss function for each video segment reads:
\begin{equation}
\nonumber
    \mathcal{L}(s)= \sum_{k \in X^+_s}\sum_{m \in X^-_s} \max(0, \mathcal{S}(v_s, f_s^k, f_o^k) -\mathcal{S}(v_s, f_s^m, f_o^m)),  
\end{equation}
where $\mathcal{S}(v_s, f_s^k, f_o^k)$ is the score output by our relation neural module in frame $k$. The score is further threshold into a binary reward $\mathcal{R}(v_s, f_s^k, f_o^k) \in \{0, 1\}$ indicating a hard decision on whether 
the visual inputs match with the natural language utterance. The threshold is predicted 
from a two-layer neural network that takes as input the relation embedding. This threshold predicting branch is trained using standard cross entropy loss for binary classification.


Natural language grounding from narrated visual demonstrations benefits from \textbf{hard negative mining} of the demonstrated natural language concepts: example frames of spatial configurations that portray the \textbf{same pair of objects, but in different spatial configurations.}  This characteristic comes for free from the way people demonstrate concepts: as suggested by psychologists \cite{doi:10.1162/1064546053278973}, related or opposite concepts are demonstrated/explained in temporal sequence, which much helps their \textbf{disentanglement}  via providing hard negative examples to the learner.  In contrast, in static images, due to photographic biases, many configurations come at stylized poses, with stylized objects,  which makes it hard for the learner to disentangle the individual characteristics of the relation. As a result, 
even with a handful of video demonstrations (14), our relation module  much improves over the pretrained model of \cite{modularreferential},  as we show  in Section \ref{sec:experiments}, while using similar unsupervised metric learning losses.   

\subsection{Policy learning with perceptual rewards}
We use the learned visual reward detectors  to train pick-and-place policies in simulation, replacing manually coded rewards, typically used in previous  works  \cite{icml2015_schulman15,hindside}.  

\paragraph{Object-factorized state representations}
Our reward detectors decompose the scoring of a spatial referential expression over an object-centric graph, where nodes represent object detections and edges  represent their spatial relationships.  
We use the same object-factorized input for our policy network and show empirically that it generalizes better than  
frame-centric representations considered in previous works \cite{DBLP:journals/corr/abs-1710-06542}, where the whole frame is provided as input to a policy network. Some recent works do also consider object-centric input
\cite{DBLP:journals/corr/abs-1708-04225,Diuk:2008:ORE:1390156.1390187}. Unlike these works, however, we additionally distinguish the roles of the objects in the scene (by mapping the subject and object in our natural language description to corresponding box hypotheses), making our object-factorized state \textbf{ordered}, as opposed to unordered.  
\paragraph{Reward shaping via analysis-by-synthesis}
Our learned reward detectors from Section \ref{sec:rewards} take as input an RGB image and a spatial natural language expression and output a binary score, $\mathcal{R}$, of whether the image matches the spatial configuration.  
Model-free policy search with binary rewards has notoriously high sample complexity  due to the lack of informative gradients for the overwhelming majority of the sampled actions \cite{NIPS2014_5421}. 
Efficient policy search  requires \textit{shaped  rewards}, either explicitly \cite{DBLP:journals/corr/LillicrapHPHETS15}, or more recently, implicitly \cite{hindside}, by encoding the goal configuration in a \textit{continuous} space where similarity can be measured against alternative goals achieved  during training. 

If we were able to visually picture  the desired 3D object  configuration to be achieved by our pick-and-place policies, then Euclidean distances to the pictured objects would provide an effective (approximate) shaping of the true rewards. We do so using analysis-by-synthesis, where our trained detector is used to select or discard  sampled hypotheses. 
Given an initial configuration of two objects that we are supposed to manipulate towards a desired configuration, we seek a physically-plausible 3D object configuration which renders to an image that scores high with our corresponding reward detector. 
Using the subject and object categories extracted from  the natural language  utterance, we retrieve 
corresponding 3D models from external 3D databases (3D Shapenet \cite{shapenet2015} and 3D Warehouse \cite{warehouse}) and import them  in a physics simulator (Bullet).  
 We sample 3D locations for the objects, render the scene and evaluate the score of our detector. 
Note that since we know the object identities, the relation module is the only one that needs to be considered for this scoring. We pick the highest scoring 3D configuration as our goal configuration. It is used at training time to provide effective shaping using 3D Euclidean distances between desired and current object locations  and drastically reduces the number of samples needed for policy learning. However, our policy network takes 2D bounding box information as input, and does not need any 3D lifting, but rather operates reactively given the RGB images. 


\section{Experiments} \label{sec:experiments}

We evaluate the accuracy of our reward detectors and their effectiveness for learning instructable  pick-and-place policies in  simulation, in place of manually coded rewards. 
Our experiments  aim to answer the following questions: 
\begin{enumerate}
\item How much does weak supervision from narrated visual demonstrations  benefit the grounding of  natural language spatial expressions, over a baseline of strongly-supervised labelled (static) image datasets? 

\item How does the accuracy of learned---as opposed to manually coded---reward detectors affect the training speed and accuracy of the corresponding reinforcement-learned policies?

\item How do object-factorized policy networks compare to their   frame-centric counter-parts? 

\item How much does reward shaping via analysis-by-synthesis help over binary rewards for efficient policy search?

\end{enumerate}

 \begin{table}[t]
\centering
\begin{tabular}{|c| c| c|c|c|c|}
 \hline
  & in & behind & left & right&avg.\\
 \hline \hline
Pretrained & 0.89 & 0.43 &0.35&0.35&0.51\\
 \hline
RandomNeg & 0.50 & 0.50 & 0.50 & 0.50 & 0.50\\
 \hline
HardNeg & {\bf 0.95} & {\bf 0.96}&{\bf 0.88}&{\bf 0.88} &{\bf 0.92}\\
 \hline
\end{tabular}
\vspace{1mm}
\caption{\textbf{Classification accuracy} of visual reward detectors of natural language spatial expressions trained in static images (\textit{pretrained}), finetuned with images using randomly selected negative examples (\textit{RandomNeg}),
finetuned with videos using hard mining negative examples (\textit{HardNeg})
for various  spatial relations. 
}
 \label{tab:classerror}
\end{table}

\begin{figure}[t!]
    \centering
    \includegraphics[width=0.85\linewidth]{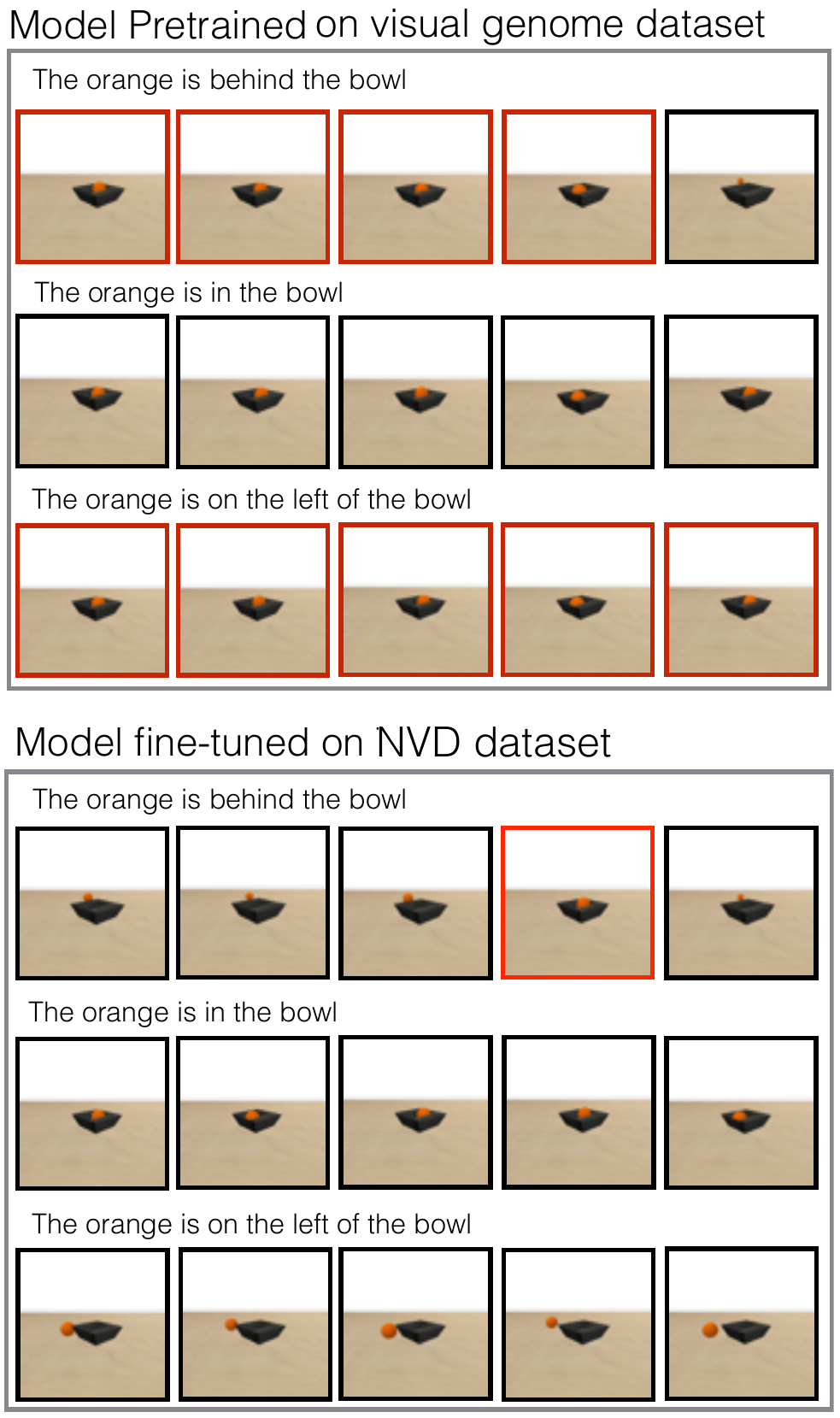}
     \centering
    \caption{ \textbf{Reward detectors} trained on static images alone (\textit{top}) and on static images and narrated video demonstrations (\textit{bottom}). We show the five highest scoring images for the two models for three spatial configurations. Red borders indicate incorrect detections. Video demonstrations improve visual detection of natural language expressions. 
    }
\label{fig:ranking}
\end{figure}

\subsection{Visual detection of natural language  expressions}
We generate a synthetic benchmark with 100 images for each spatial relationship.
The relationships we consider are \textit{in, behind, left, and right}.
Each set of 100 images has 50 positive and 50 negative images. 
Ground truth annotations are generated by a hard-coded function
in the simulator.
In Table \ref{tab:classerror}, we show classification error of the learned visual detectors.  We compare the  reward detector described in Section \ref{sec:rewards} trained on Visual Genome  and finetuned on the video demonstration dataset, against the  network of Hu et al. \cite{modularreferential} trained on Visual Genome  \cite{krishnavisualgenome}.  

In Figure~\ref{fig:ranking}, we show the top retrieved images in a pool of 75 images that depict diverse spatial configurations of the same two objects (orange and bowl) using the (unthresholded) scores of our detectors.  
In both the classification task and the retrieval task, finetuning in our small video dataset helps the detector, despite using only 14  demonstration videos.

Finetuning in our NVD dataset clearly improves upon the pretrained model. 
Our video demonstrations often show \textbf{multiple spatial configurations of the same pair of objects}, and the data therefore have less biases regarding configuration-category correlations than static images.
We further compare the hard negative mining from our NVD dataset  against random 
sampling for negative examples from 
Visual Genome  \cite{krishnavisualgenome} in Table \ref{tab:classerror}.  
Hard negative mining in NVD helps over random negative examples  from the static image dataset (random in the absence of any  information for sampling more informative negative examples). 

\begin{figure}[t!]
    \centering
    \includegraphics[width=1.0\linewidth]{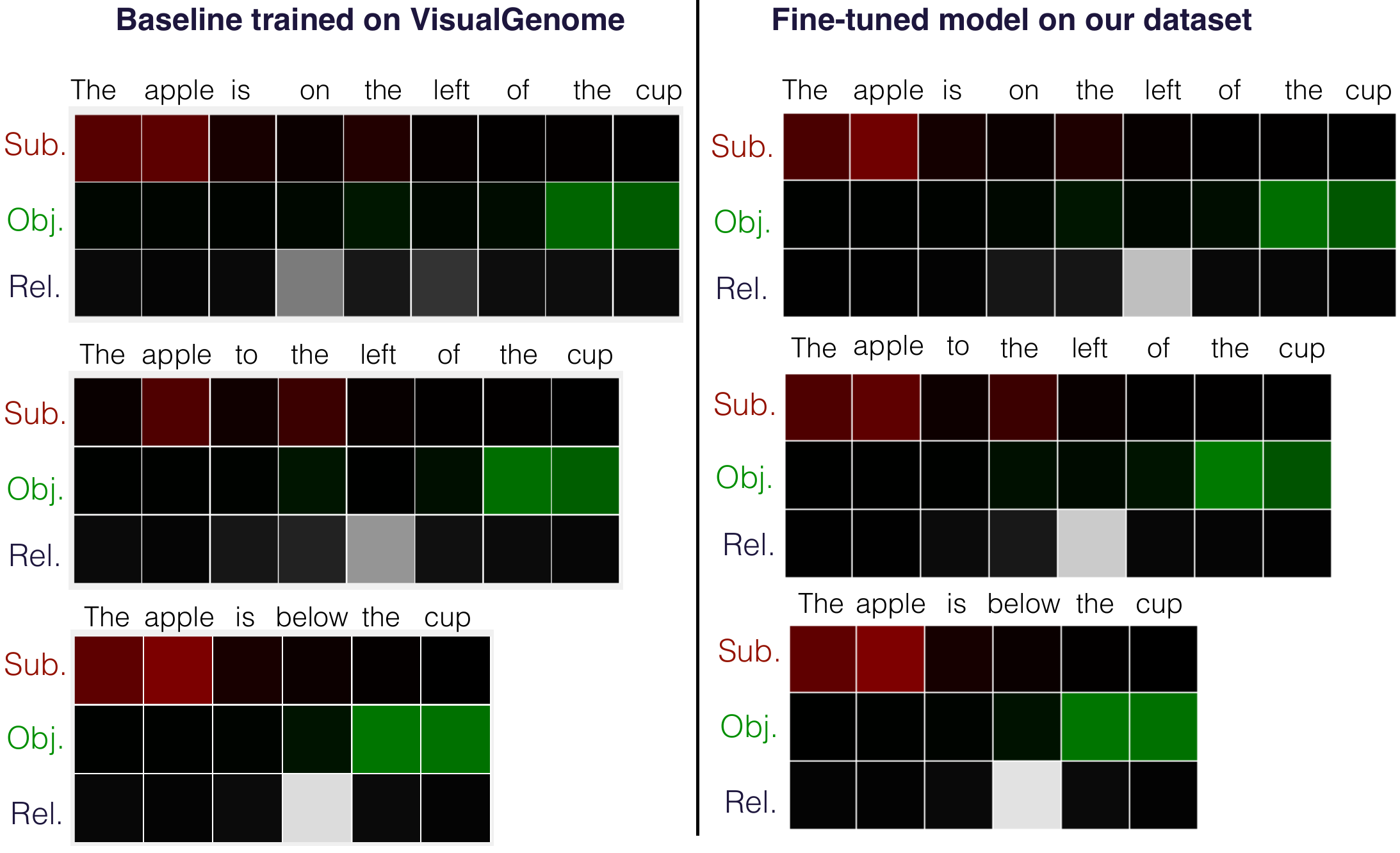}
     \centering
     \vspace{-2mm}
    \caption{ \textbf{BiLSTM attention weights on language representation} on unseen natural language  descriptions. The  detector trained from video demonstrations places weight on more informative keywords and generalizes to unseen sentences. 
    }
\label{fig:relation}
\end{figure}

\begin{figure}[h!]
    \centering
     \includegraphics[width=1.0\linewidth]{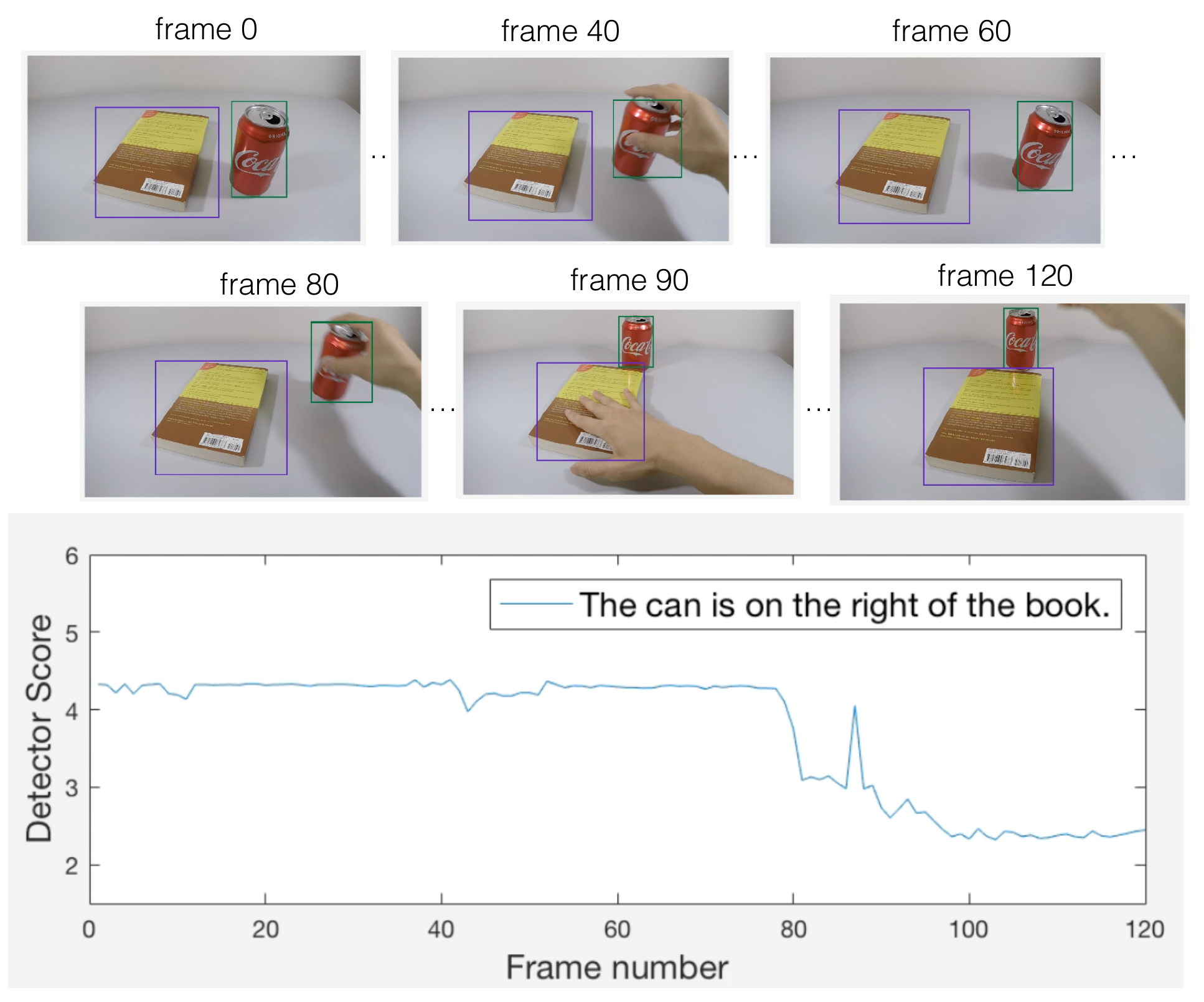}
     \centering
    \caption{ \textbf{Reward detection on real test videos}. 
    }
\label{fig:orange}
\end{figure}

In Figure \ref{fig:relation}, we visualize BiLSTM attention weights over the hidden states of the language representation from the pretrained and finetuned model. The finetuned model is placing weights on more informative keywords for relations, e.g., ``right" and ``left", and is able to generalize 
to unseen (novel) natural language descriptions. Despite the fact that our model does not use the word embedding of the object or the subject, those also improve through the gradients on the relationship. In Figure \ref{fig:orange}, we show detector scores on real video sequences.


\subsection{Policy learning with Perceptual Reward Detectors}
We use our learned detectors to train instructable pick-and-place  policies 
in the Bullet physics simulator \cite{bullet}. 
Our policy always starts by grasping the subject of the natural language utterance as detected by our object detector.  We use deep Q learning \cite{DBLP:journals/corr/MnihKSGAWR13} over a discrete action set of \{`move forward,' `move backward,' `move right,' `move left'\} to learn a model-free policy that moves the end-effector of the Kuka IIWA robotic arm so that after an episode length of $T = 5$ action steps, the gripper opens, the object is released, and the desired configuration is achieved. 
Our policy network is a convolutional neural network  that takes an RGB frame as input and produces a distribution over our action set. We will call this policy network \RGBnet, to distinguish it from \StructureNet, which takes the spatial configuration of two objects, instead of the RGB.

\paragraph{Implementation details} 
\RGBnet \ has 5 convolutional layers and 3 fully connected layers with filter size $5\times5$ (stride 2), $5\times5$ (stride 2), $3\times3$ (stride 1), $5\times5$ (stride 2), and $3\times3$ (stride 1), respectively. Channel sizes are set to 32, 32, 32, 16, and 16, respectively. We use ReLU as the activation function. To reduce memory usage, we shrink the input image to $72 \times 72.$ The \ \StructureNet \ consists of three fully connected layers with size of $512$, $512,$ and $4.$   We use ReLU as activation function after the first two layers. We train both networks starting from random weights using the Adam optimizer and learning rate of 0.001. The batch size in both models is set to 512. In each episode (trial), with exploration rate $\alpha$, DQN takes
randomly-selected actions with probability $\alpha$ and the action with highest score with probability $(1-\alpha).$  
The exploration rate for DQN training is set to 0.8 and decays with the rate of 0.1 every 1000 action selections. 
For every five action selections, we take one gradient descent step for the DQN.

\paragraph{Task} The task is putting an object inside a container. 
The containers are always facing up and initialized in
a $15cm$ $\times$ $40cm$ region with randomly selected orientation. The size of the container is roughly $5cm$ $\times $ $5cm$. The subject indicated by our parser  is initialized to be grasped by the gripper
and hanging $20cm$ above the table. 


\paragraph{Reward shaping} We found that   binary (oracle) rewards were not able to train successful policies with episode length larger than $T=2$. 
When shaped rewards are combined with binary rewards, in terms of Euclidean distance between current and desired object 3D locations,  effective policies were learned even when starting far away from the desired end-effector position. 
Thus, all our policy learning results in this section are obtained by combining (i) oracle shaping with oracle binary rewards ($R_{gt}$), or (ii) predicted shaping using analysis-by-synthesis with predicted binary rewards from our learned reward detectors  ($R_d$).

\begin{figure}[t!]
    \centering
    \includegraphics[width=1.0\linewidth]{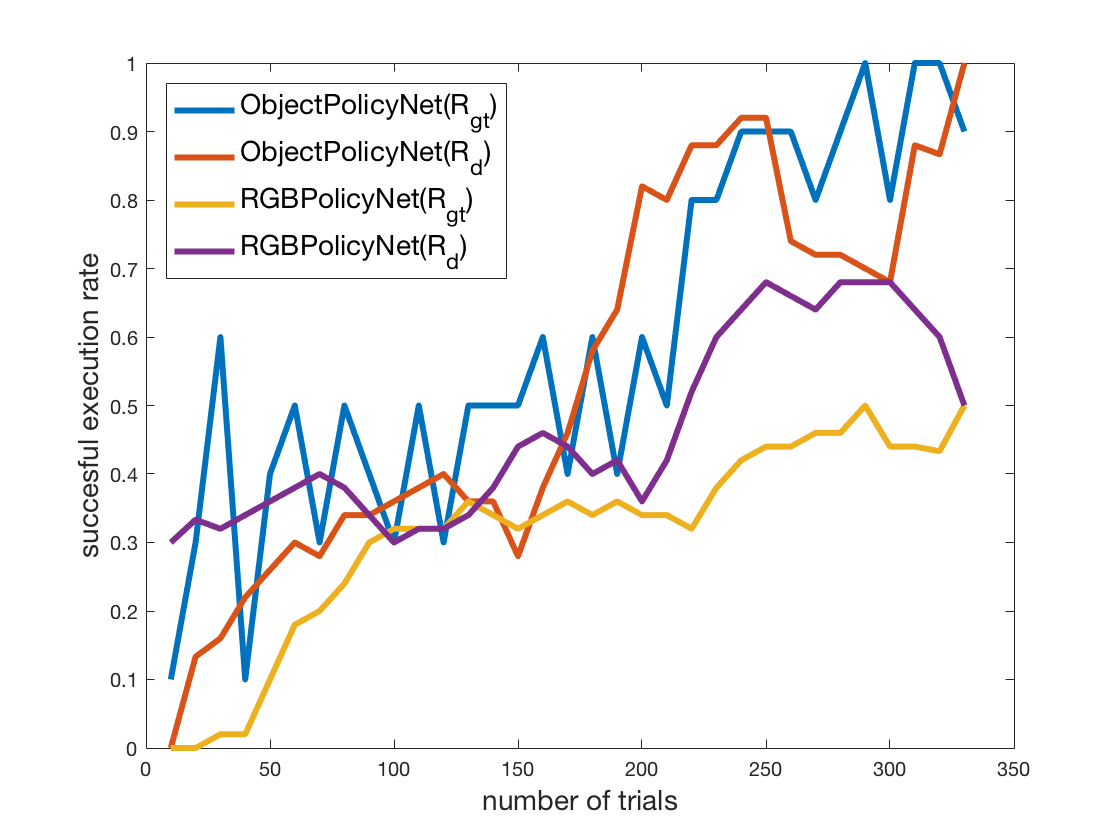}
     \centering
    \caption{ \textbf{Policy learning with/wo noisy rewards, with/wo object-factorized  input.}
    }
\label{fig:curve}
\end{figure}

\paragraph{Noisy  rewards}
We show in Figure \ref{fig:curve} plots of test  policy accuracy against the number of episodes for \RGBnet \ and \StructureNet \ using (i) oracle rewards ($R_{gt}$), and (ii) learned rewards ($R_d$) for the instruction ``put the orange inside the bowl" . In a synthetic dataset of balanced successful and unsuccessful configurations, our reward detector has a classification accuracy of 95\%. Table \ref{tab:results1} shows that policy learning from noisy visual rewards for \StructureNet \ has 8\% lower training accuracy, and much lower test performance than  a policy trained with oracle rewards. 

\RGBnet \ is not strongly affected by whether the rewards are provided by an oracle or predicted by perception. 

\paragraph{Object-factorized state representations} 
In Figure \ref{fig:curve} and in Table \ref{tab:results1}, we compare \RGBnet \ and \StructureNet \ in their performance on seen and unseen objects. \RGBnet \
 does considerably worse, especially on unseen objects. \RGBnet \ does 
 not have a way to generalize to new object appearances at test time. Its worse performance during training can be explained as underfitting. It is severely hurt by resolution,   
since we wildly vary the configuration of the two objects during training. 



\begin{table}[t!]
\centering
\begin{tabular}{|c| c| c|}
 \hline
 & \textbf{accuracy} & \textbf{accuracy}\\
  & (seen objects) & (unseen objects)\\
 \hline \hline
 \StructureNet($R_{gt}$) & 0.96 & 0.78 \\
 \hline
 \StructureNet($R_{d}$) & 0.88 & 0.50\\
 \hline
 \RGBnet($R_{gt}$) & 0.71 & 0.27 \\
 \hline
  \RGBnet($R_{d}$) & 0.71 & 0.40\\
 \hline
\end{tabular}
\vspace{1mm}
\caption{\textbf{Train/test policy accuracy (\% of successful trials) for learning the task ``place objects inside containers.''} We consider two different policy network structures: (i) object bounding boxes and their spatial features  as input  (\StructureNet), and (ii) RGB image as input (\RGBnet). We compare  policies learned by manually-coded rewards
 in the simulator ($R_{gt}$) and by
our learned reward detector  ($R_{d}$). We compare policies on 
objects seen during training (but in novel positions), and on  novel objects. 
}
 \label{tab:results1}
\end{table}


\section{Conclusion}

In this work we introduce
a paradigm for learning instructable pick-and-place policies through reinforcement from perceptual reward detectors trained through grounding narrations in narrated visual demonstrations. 
We show how the accuracy of the reward detectors affects the accuracy of the learned policies, and how   object-factorized state representations that follow the syntactic structure of natural language  help  generalization of rewards and policies to novel scenes. We further show how goals instructed in natural language allow the description of novel goals and programming of corresponding novel behaviours at test time. 
Future work involves scaling up the vocabulary acquired for describing goals of activities, and also the corresponding skill library. Additionally, the training currently done in simulation can be done on a robotic platform. Finally, we plan to use more of the narrated demonstrations, rather than merely the final goal configurations.  

\paragraph{Acknowledgements}
The authors would like to thank Hsiao-Wei Tung, 
Samuel Pepose,
Ishu Garg, Medha Potluri, and
Kanthashree Mysore Sathyendra
for contributing to the NVD Dataset.

{\small
\bibliographystyle{ieee}
\bibliography{egbib}
}

\end{document}